\documentclass[runningheads]{llncs}
\usepackage[T1]{fontenc}
\usepackage{graphicx}
\usepackage{hyperref}
\usepackage{algorithm}
\usepackage{algpseudocode}
\usepackage{array}

\begin{document}
\title{Derivation Prompting: A Logic-Based Method for Improving Retrieval-Augmented Generation}

\titlerunning{Derivation Prompting: A Logic-Based Method for Improving RAG}

\author{Ignacio Sastre \and
Guillermo Moncecchi \and
Aiala Rosá}

\authorrunning{I. Sastre et al.}

\institute{Instituto de Computación, Facultad de Ingeniería, Universidad de la República\\Montevideo, Uruguay\\
\email{\{isastre,gmonce,aialar\}@fing.edu.uy}}

\maketitle

\begin{abstract}

The application of Large Language Models to Question Answering has shown great promise, but important challenges such as hallucinations and erroneous reasoning arise when using these models, particularly in knowledge-intensive, domain-specific tasks.
To address these issues, we introduce Derivation Prompting, a novel prompting technique for the generation step of the Retrieval-Augmented Generation framework.
Inspired by logic derivations, this method involves deriving conclusions from initial hypotheses through the systematic application of predefined rules. It constructs a derivation tree that is interpretable and adds control over the generation process.
We applied this method in a specific case study, significantly reducing unacceptable answers compared to traditional RAG and long-context window methods.\footnote{Repo with all prompts: \url{https://github.com/nsuruguay05/derivation-prompting}}

\keywords{Large Language Models \and Retrieval-Augmented Generation \and Question Answering}
\end{abstract}

\section{Introduction}

Question Answering (QA) has improved substantially with the advent of Large Language Models (LLMs).
However, these models face important challenges, particularly in knowledge-intensive, domain-specific tasks, such as hallucinations and faulty reasoning~\cite{valmeekam2022large,ji2023hallucination}.
The Retrieval-Augmented Generation (RAG) framework addresses these limitations by retrieving the most relevant document chunks from a trusted domain-specific document base, and grounding the LLM’s generation on the retrieved information~\cite{gao2024retrievalaugmented}.

\begin{figure}
\centering
\includegraphics[width=\textwidth]{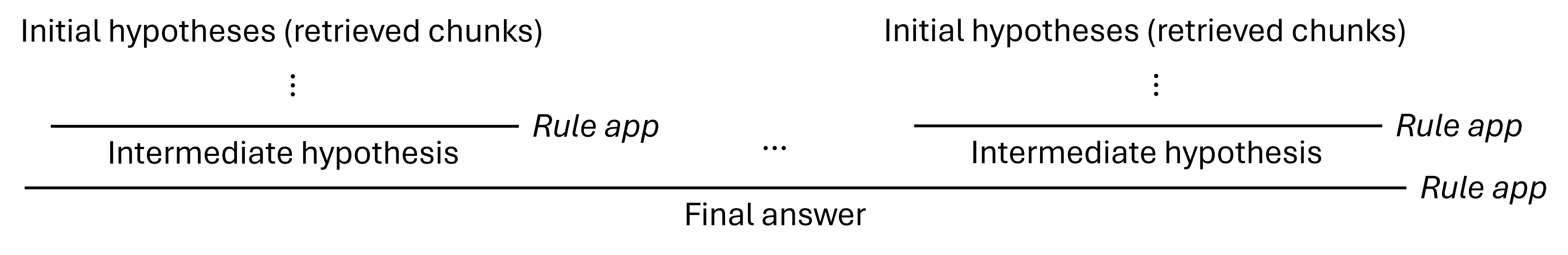}
\vspace{-0.4cm}
\caption{Schematic illustration of a derivation tree constructed using derivation prompting.} \label{fig:derivation-prompting}
\end{figure}

Substantial work has been done to improve the reasoning ability of LLMs~\cite{huang-chang-2023-towards}.
Techniques like Chain-of-Thought (CoT)~\cite{wei2023chainofthought} have reliably improved performance across various tasks, including QA.
However, these techniques do not explicitly define \textit{how} the model should reason, as there are no restrictions on how the intermediate reasoning steps should be constructed.

In this work, we propose Derivation Prompting, an alternative approach for the generation step in the RAG framework inspired by logical derivations.
In this method, a conclusion is inferred from initial hypotheses by applying well-defined rules to transform and/or combine these hypotheses.
This novel approach offers some advantages over existing methods, mainly:

\begin{itemize}
    \item \textbf{Interpretability:} The method not only generates a final answer but also produces a tree structure, referred to as a derivation (see Figure~\ref{fig:derivation-prompting}).
    Each node in the derivation represents the application of an easily interpretable rule that transforms some of its children.
    This structure provides a straightforward way to identify errors the model could have made and to understand how it arrived at the final answer.
    
    \item \textbf{Controlled generation:} By generating the answer through the sequential application of predefined rules, this method provides a clearer reasoning path for the model to follow.
    This reduces hallucinations and faulty reasoning while ensuring that the generated answers remain grounded in the information from the documents.
\end{itemize}

The paper is structured as follows:
Section~\ref{sec:related-work} presents related work.
Section~\ref{sec:derivation} provides a detailed explanation of Derivation Prompting.
Section~\ref{sec:case-study} describes the case study conducted.
Section~\ref{sec:evaluation} explains the evaluation method used.
Section~\ref{sec:results} presents the results and analysis.
Finally, Section~\ref{sec:conclusions} contains the conclusions and outlines future work.

\section{Related work}
\label{sec:related-work}

\noindent\textbf{Retrieval Augmented Generation}

Derivation Prompting, as proposed in this work, is a prompting technique applied within the context of Retrieval-Augmented Generation (RAG).
RAG augments LLMs with a retrieval component that recovers the most relevant information from an external knowledge base~\cite{lewis_retrieval-augmented_2020,izacard2024atlas,shi2023replug}.

The naive RAG paradigm consists of two main steps: retrieval and generation.
First, documents are segmented into smaller units, referred to as chunks, which are subsequently converted into vector representations.
Upon a user query, during the retrieval step the query is converted to its vector representation, and similarity scores between the query and the indexed chunks are computed.
The top $k$ chunks are retrieved and used as context in the prompt for the generation step~\cite{gao2024retrievalaugmented}.

An alternative to using vector representations is to employ Cross-Encoder models that directly process each chunk with the query and return a similarity score~\cite{nogueira2020passage,reimers_sentence-bert_2019}.
While this approach tends to yield better results, it is significantly less compute-efficient, as it requires as many model inferences per query as number of chucks we have, compared to only one inference when using sentence embeddings.\\

\noindent\textbf{Prompting techniques for enhancing reasoning}

Substantial work has been done to improve the reasoning ability of LLMs~\cite{huang-chang-2023-towards}.
Chain-of-Thought (CoT) \cite{wei2023chainofthought} involves prompting the model to generate a coherent series of intermediate reasoning steps that lead to the final answer.
Few-Shot prompting~\cite{brown2020language} is applied and a chain-of-thought is added to each example.
They show that sufficiently large LLMs can generate these reasoning chains, yielding promising results in arithmetic, commonsense, and symbolic reasoning tasks.

The Tree of Thoughts (ToT) framework~\cite{yao2023tree} is an evolution of CoT that enables models to explore several different reasoning paths.
In this approach, reasoning is conceptualized as searching through a tree, where each node represents a thought.
This framework addresses some limitations of CoT, such as the inability to explore different continuations within the same reasoning chain or to backtrack when incorrect conclusions are reached.

While these methods significantly improve performance on various tasks, there is no control over how each thought is generated in the chain, as there is no systematic methodology the model has to follow.
This lack of control can lead to erroneous reasoning and susceptibility to hallucinations, which are well-known problems when working with LLMs~\cite{valmeekam2022large,ji2023hallucination}.\\

\noindent\textbf{Logic and LLMs}

Similar to Derivation Prompting, some works explore combining classical logic with prompting techniques to improve reasoning.
The Logical Thoughts (LoT) prompting framework~\cite{zhao2024enhancing} uses logical equivalence, expressing premises in various logically equivalent forms to encourage the exploration of different solutions.
This is achieved by incorporating a verification step for each thought, where an explanation is generated for both the thought as-is and its logical negation.
The LLM is then tasked to decide between the two.

Symbolic CoT (SymbCoT)~\cite{xu2024faithful} is another proposed method that involves four LLM modules:
(i) Translator: translating premises and the question to First-Order Logic formulas,
(ii) Planner: dividing the original problem into smaller subproblems and developing a step-by-step plan,
(iii) Solver: deriving the answer through a logical inference process and
(iv) Verifier: validating the correctness of the translations and the Solver’s output.

\section{Derivation prompting}
\label{sec:derivation}

This technique focuses on the generation step of the Retrieval-Augmented Generation (RAG) framework.
It relies on the premise that the expected answer to a given query must be obtained by combining and/or transforming the most relevant information extracted from a document base, since our objective is to rely only on the information available in the documents and not on the information the model could have learned on its training phase.

\begin{figure}
\centering
\includegraphics[width=.4\textwidth]{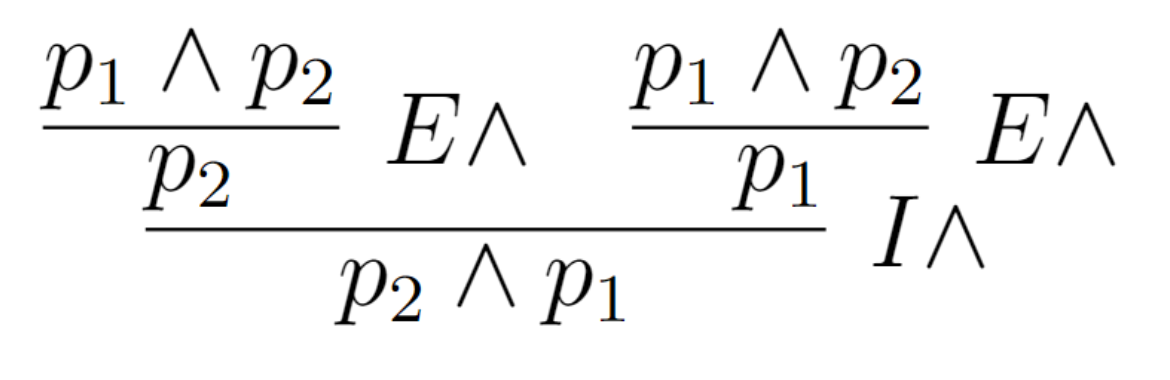}
\vspace{-0.4cm}
\caption{Example of a derivation proving the statement $p_1 \wedge p_2 \vdash p_2 \wedge p_1$, where $p_1$ and $p_2$ are proposition symbols and $E\wedge$ and $I\wedge$ are the elimination and introduction rules respectively for $\wedge$, as defined in \cite{van_dalen_logic_2013}.}
\label{fig:derivation-example}
\end{figure}

The idea for this technique is inspired by how a derivation tree in propositional logic is constructed.
In this context, a conclusion $\varphi$ is derived from a set of premises or hypotheses $\Gamma = \{\delta_1, \dots, \delta_n\}$.
We denote $\Gamma \vdash \varphi$ if such a derivation exists.
The class of derivations forms an inductively defined set characterized by a list of inference rules that explicitly state how to derive new conclusions from existing ones.
These rules operate by systematically applying logical operations to the premises to construct a tree where each node represents an application of a rule, culminating in the conclusion at the root~\cite{van_dalen_logic_2013}. Figure~\ref{fig:derivation-example} shows an example of a logic derivation.

\vspace{-0.4cm}

\vspace{-0.4cm}

In a typical RAG framework, documents are divided into smaller units called chunks.
Given a query, the $n$ most relevant chunks are selected and used as context for generating the answer. Following the analogy with logic derivations, in Derivation Prompting, we consider these most relevant chunks as a set of hypotheses $\{h_1, \dots, h_n\}$.
The objective is to construct a derivation tree using predefined natural language rules, ultimately deriving a conclusion $c$, such that $h_1, \dots, h_n \vdash c$, as depicted in Figure~\ref{fig:derivation-prompting}.

In contrast to logic derivations, where we usually start from a candidate conclusion and seek to construct the proof, in this case, the conclusion is not known beforehand.
Therefore, a query $q$ is needed to guide the construction of the derivation tree, with the goal that the resulting conclusion serves as the answer to the query $q$.

For each step in the construction of the derivation, the task the LLM has to follow involves deciding which rule to apply, selecting the appropriate hypotheses, and generating the conclusion that arises from the application of the chosen rule.
Although it may seem counter-intuitive to let the LLM decide which rule to apply, this is a key aspect for making this method viable due to the LLM's ability to disambiguate natural language in both the rule explanation and the hypotheses.

\subsection{Rules}

\begin{figure}[H]
\centering
\includegraphics[width=\textwidth]{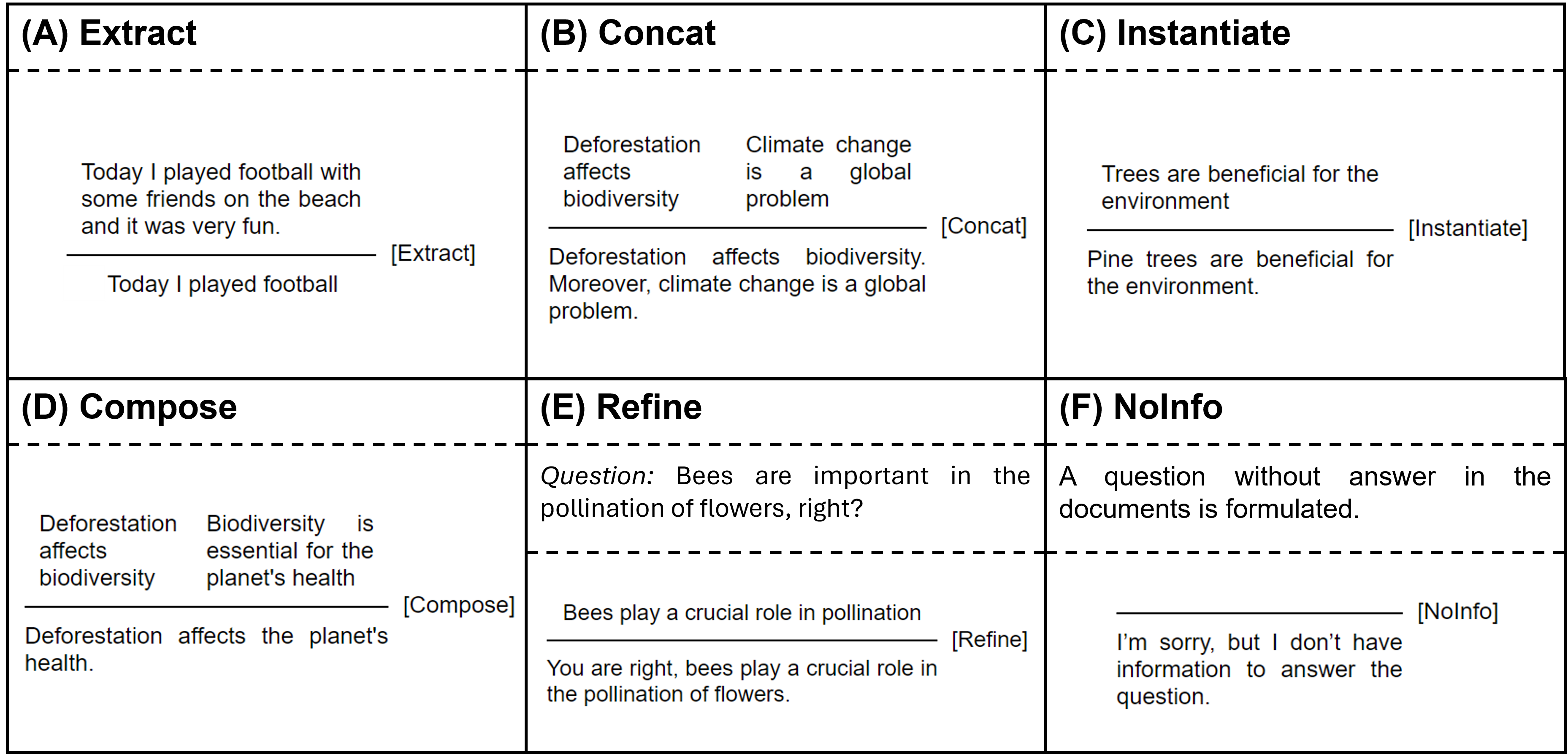}
\vspace{-0.4cm}
\caption{Toy examples of application for each rule. Examples (E) and (F) have information of the query for better understanding.} \label{fig:rules-examples}
\end{figure}

For Derivation Prompting, a set of derivation rules must be defined.
These rules are specified in natural language and used by the language model to construct a derivation tree.
We define a set of derivation rules that are convenient for our use case (Section~\ref{sec:case-study}).
It is important to notice that these rules are specific to this problem, and any set of rules could be defined that best fits the type of combinations and/or transformations necessary in different use cases.
Table~\ref{tab:rules} presents each rule with a description and figure~\ref{fig:rules-examples} shows a toy example for each rule.

\begin{table}
\caption{List of defined rules with a brief description.}
\label{tab:rules}
\begin{tabular}{|l|p{10.2cm}|}
\hline
\textbf{Name} & \textbf{Description}\\
\hline
Extract &  Given a hypothesis $h$, this rule extracts a specific part of $h$ as a conclusion.\\
\hline
Concat &  Combines two independent hypotheses to generate the conclusion.\\
\hline
Instantiate & Generates a conclusion by instantiating a generic hypothesis into a particular case.\\
\hline
Compose & Combines two hypotheses that share a common element to generate a new conclusion.\\
\hline
Refine & Given a hypothesis $h$, it slightly adapts it to better fit the question, without modifying the semantics or content of $h$.\\
\hline
NoInfo & This rule is used when none of the hypotheses provide information to answer the question (or part of the question).\\
\hline
\end{tabular}
\end{table}

\vspace{-0.8cm}

\subsection{Algorithm}

Algorithm~\ref{alg:derivation-prompting} presents the pseudo-code for constructing a derivation.
When looking at the algorithm in detail, it is important to notice that lines 3, 4, and 5 correspond to steps that the LLM should execute.
The responsibilities of the LLM are to decide which rule to apply and which hypotheses to use, as well as to construct the conclusion that arises from the application of the chosen rule.
Additionally, the LLM is used to determine whether the conclusion serves as the final answer to the user's query.

\vspace{-0.4cm}

\begin{algorithm}[h!]
\caption{Derivation Prompting pseudo-code}
\label{alg:derivation-prompting}
\begin{algorithmic}[1]
\Require $hypotheses\_list = \{h_1, \dots, h_n\}$: List of hypotheses, $q$: Query
\State $final\_answer \gets \textbf{False}$
\While{\textbf{not} $final\_answer$}
    \State Decide which rule $r$ to apply
    \State Decide which hypotheses $\{h_i, \dots, h_k\}$ to apply $r$ to
    \State $conclusion \gets$ apply rule $r$ over $\{h_i, \dots, h_k\}$ and query $q$
    \If{$conclusion$ is the final answer}
        \State $final\_answer \gets \textbf{True}$
    \Else
        \State $hypotheses\_list.append(conclusion)$
    \EndIf
\EndWhile
\State \Return $conclusion$
\end{algorithmic}
\end{algorithm}

\vspace{-0.4cm}

The rest of the algorithm is straightforward.
If the conclusion is considered the final answer, the derivation is complete, and the last conclusion is used as the answer.
If not, the conclusion is added to the list of hypotheses and can be used in subsequent rule applications (though it might potentially never be used).
Optionally, each rule application can be stored with pointers to its arguments and conclusion to later reconstruct the derivation tree.

We explored different ways of implementing the aforementioned algorithm and considered two main alternatives:

\begin{enumerate}
    \item \textbf{One-step prompt:} This approach isolates each rule application as an independent LLM call.
    Given the list of hypotheses in the middle of a derivation, the model is prompted to produce, in a single inference, the rule to apply, the hypotheses to use, the resulting conclusion, and whether it is the final answer.
    The algorithm is implemented similarly to the Algorithm~\ref{alg:derivation-prompting}, with lines 3, 4, and 5 replaced by this single call to the LLM, followed by parsing the result.

    \item \textbf{Whole derivation prompt:} In contrast to the previous alternative, this approach allows the LLM to construct the entire derivation in one inference call, effectively emulating the execution of Algorithm~\ref{alg:derivation-prompting}.
    To achieve this, we applied a Few-Shot strategy~\cite{brown2020language}, crafting six complete examples of manual executions of the algorithm to create different derivations using all the rules (Appendix~\ref{sec:few-shot} shows one of these examples).
    The model is then prompted to follow the same steps with a new query and initial hypotheses.
    The result is then parsed to obtain each rule execution and intermediate hypotheses.
\end{enumerate}

In our experiments, the second approach yielded results as good as the first one but was much faster and computationally cheaper, as it replaces $n$ LLM inferences with just one. Therefore, we decided to extensively use the whole derivation approach.

\section{Case study}
\label{sec:case-study}

We investigated this idea in the context of a specific use case: developing a platform for question answering in the domain of administrative information for the School of Engineering at Universidad de la República (UDELAR), for the Spanish language.
Currently, the school operates the Orientation and Consultation Space (OCS), where students can ask questions via email or in person, and OCS staff provide answers.
We explored the feasibility of building a tool to assist with this work by offering an automated system for students to obtain answers to their questions.

We gathered a small set of documents available on the school’s webpage.
Specifically, 17 websites were scraped and converted to markdown format using LangChain’s \texttt{Html2TextTransformer} class.

For evaluation purposes, we constructed a QA dataset consisting of 135 real user queries.
These queries were derived from past emails sent to the OCS over the last few years.
Each student email was preprocessed using the Llama 2 7B model~\cite{touvron_llama_2023} to remove irrelevant information typical of email communication (e.g., greetings, apologies) and, most importantly, personal information such as names, identification numbers, and phone numbers.

In the context of this project, we explored several methods using LLMs which are explained below:

\noindent\textbf{Retrieval Augmented Generation (RAG)}

For the retrieval step, we explored using sentence embeddings generated with \texttt{intfloat/multilingual-e5-large} as well as \texttt{BAAI/bge-reranker-large} Cross-Encoder model.
Given that the use case involves fewer than a hundred chunks, using a Cross-Encoder was feasible and, as expected, consistently yielded better results than using sentence embeddings.

For the generation step, we experimented with models from the Anthropic’s Claude 3 family~\footnote{\url{https://www.anthropic.com/news/claude-3-family}} (specifically, Haiku, which is faster but less capable, and Opus, which is the best performing and competitive with OpenAI’s GPT-4).
The $k$ most relevant chunks to the user’s query were added as context, and a prompt was crafted to explicitly instruct the model to use them for generating the answer.

\noindent\textbf{Using Long Context Windows}

Another approach we explored was leveraging the long context windows of closed models, specifically the Claude models, which support up to 200k tokens.
In this method, we inserted all the full documents as context, thereby avoiding the retrieval step.

\noindent\textbf{Derivation Prompting}

Utilizing the retrieval method described in the RAG experiment, we explored the use of Derivation Prompting for the generation step, as detailed in Section \ref{sec:derivation}.
For this method, we used three initial hypotheses, corresponding to the three most relevant chunks obtained in the retrieval step.
We did not explore using more hypotheses because of how the Few-Shot examples were designed, but this is an area for future work that we plan to investigate.

\section{Evaluation}
\label{sec:evaluation}

Evaluating Open-Domain Question Answering, especially when using LLMs, remains an open problem, and human evaluation still appears to have no substitute~\cite{kamalloo-etal-2023-evaluating}.
Nevertheless, it has been shown that state-of-the-art LLMs tend to exhibit a high degree of agreement with human evaluation when used as judges~\cite{zheng2023llmjudge}.
Therefore, we decided to follow this approach for evaluating each experiment separately.
We are currently conducting human evaluation on the best performing experiments and results will be presented in future work.

We designed an evaluation prompt following the format defined for the Feedback Collection dataset~\cite{kim2024prometheus}, which encompasses four components:

\begin{enumerate}
    \item \textbf{Instruction to evaluate:} The instruction for the task to evaluate. In our case, this is the particular question that the answer addresses. 
    \item \textbf{Response to Evaluate:} The response to the question that the LLM has to evaluate (with a score on a scale from 1 to 5).
    \item \textbf{Reference Answer:} A reference answer that corresponds to a score of 5.
    \item \textbf{Customized Score Rubric:} Specific criteria defined for our use case, specifying what the evaluator should focus on. This includes a description of the criteria and a detailed explanation for each possible score (1 to 5).
\end{enumerate}

We defined the score rubric criteria as determining whether the generated answer is correct and truthful.
This is clearly specified in each score description. Table~\ref{ref:score-rubric} presents a brief explanation of each score. 

\begin{table}
\caption{Score rubric criteria defined for each score, for evaluating generated answers.}
\label{ref:score-rubric}
\begin{tabular}{|c|p{8.5cm}|c|}
\hline
\textbf{Score} & \textbf{Explanation} & \textbf{Classification}\\
\hline
1 & Candidate contradicts reference; false information. & Unacceptable \\
\hline
2 & Candidate has conflicts with reference; partially false information. & Unacceptable \\
\hline
3 & Candidate does not contradict reference but does not provide any information either. & Acceptable \\
\hline
4 & Candidate partially matches reference; correct but incomplete information. & Acceptable \\
\hline
5 & Candidate completely matches reference; correct and complete information. & Acceptable \\
\hline
\end{tabular}
\end{table}

Additionally, we classified scores 1 and 2 as unacceptable and scores 3 to 5 as acceptable, thereby obtaining an aggregated metric for evaluation.
Scores 1 and 2 correspond to answers that fully or partially contradict the reference answer and are therefore considered unacceptable.
Scores 3 to 5 may have none of the information correct (but are not incorrect either, as no information is provided at all), part of the information correct, or be fully correct.
In all these cases, the answers do not provide false or contradictory information and are considered acceptable.

\section{Results}
\label{sec:results} 

The evaluation was carried out using Claude Opus as the evaluator.
Table~\ref{table:results} shows the percentage of acceptable answers and the distribution for each score from 1 to 5 for the best performing experiments.
These experiments utilize Claude Opus and Claude Haiku, and where applicable, the Cross-Encoder for the retrieval step, with the number of chunks used as context set to $k = 3$.

\begin{table}[h]
\caption{Percentage of acceptable answers, distribution of scores, average and standard deviation metrics for each experiment. CH is Claude Haiku and CO is Claude Opus.}
\label{table:results}
\centering
\begin{tabular}{|c|c|c|c|c|c|c|c|c|}
\hline
\textbf{Experiment} & \textbf{\% Accep.} & \textbf{\#1} & \textbf{\#2} & \textbf{\#3} & \textbf{\#4} & \textbf{\#5} & \textbf{Avg} & \textbf{Std Dev} \\
\hline
Long context - CH  & 65.2 & 35 & 12 & 61 & 22 & \textbf{5} & 2.63 & 1.14 \\
RAG - CH  & 72.6 & 17 & 20 & 74 & 20 & 4 & 2.81 & 0.94 \\
Derivation Prompting - CH & 82.2 & 21 & 3 & 95 & 12 & 4 & 2.81 & 0.91 \\
Long context - CO & 76.3 & 17 & 15 & 73 & 25 & \textbf{5} & 2.90 & 0.97 \\
RAG - CO & 77.8 & 15 & 15 & 75 & 27 & 3 & 2.91 & 0.92 \\
Derivation Prompting - CO & \textbf{89.6} & \textbf{10} & 4 & 92 & 25 & 4 & \textbf{3.07} & \textbf{0.79} \\
\hline
\end{tabular}
\end{table}

As can be observed in Table~\ref{table:results}, Derivation Prompting with Claude Opus significantly reduces the number of unacceptable answers compared to the other experiments.
However, it does not necessarily increase the number of answers with scores of 4 and 5.
Many unacceptable answers from the other experiments receive a score of 3 in Derivation Prompting.
There are two primary reasons for this:
(1) The NoInfo rule has a more direct impact than simply prompting the model not to answer questions when the information is not available, as done in both the RAG and Long Context experiments;
(2) Generating answers through the application of explicitly defined and constrained rules reduces the likelihood of hallucinations or misinterpretations of the context chunks and minimizes the potential for faulty reasoning.
These results suggest that while there is minimal impact on recall (i.e., answering as much as possible), there is a significant  improvement in precision (i.e., avoiding incorrect answers).

It is important to note that while Derivation Prompting with Claude Haiku also reduces unacceptable answers, it does have an impact on the number of answers with scores of 4 and 5, resulting in fewer such answers compared to RAG.
This suggests that the size of the model is an important factor.
Larger and more powerful models, such as Claude Opus, have a better understanding of the task of constructing the derivation and are more capable of applying the rules effectively, yielding better results.

\begin{figure}
\centering
\includegraphics[width=\textwidth]{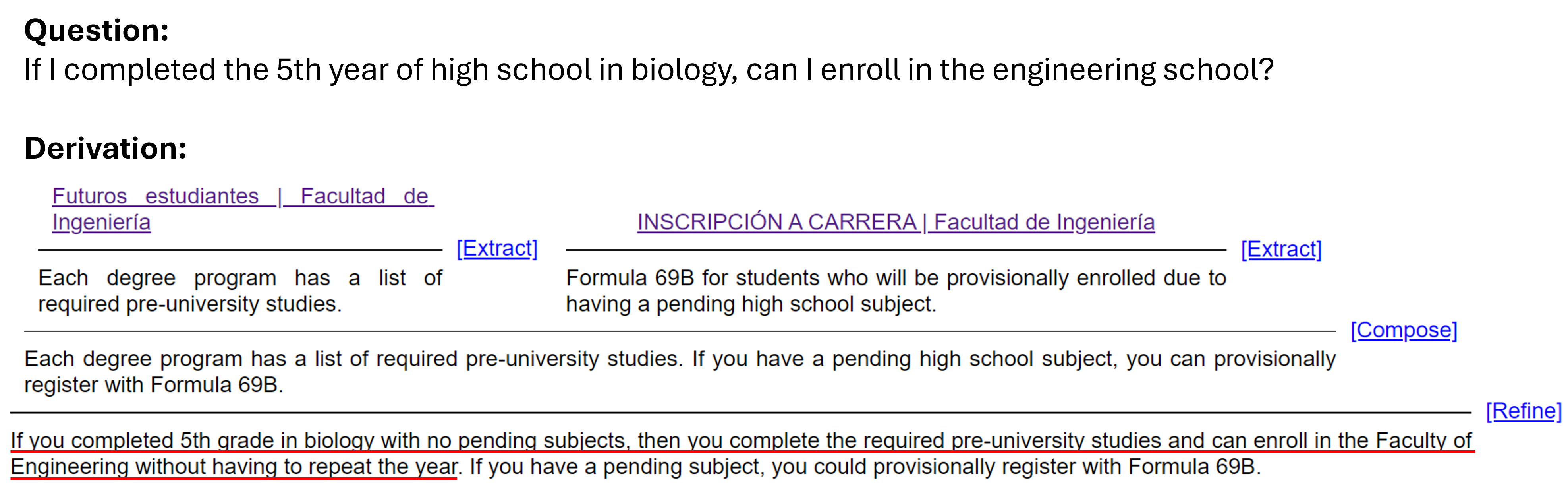}
\vspace{-0.4cm}
\caption{Example of an incorrect derivation (translated from Spanish). In the application of the Refine rule, the model hallucinates that having completed 5th year of high school in biology fulfills the required pre-university studies (hallucination is underlined in red).} \label{fig:incorrect-example}
\end{figure}

Although unacceptable answers have been reduced, there are still some examples that scored 1 and 2.
A significant advantage of Derivation Prompting is that the resulting derivation is interpretable, and it is easy to identify mistakes in the application of rules.
This, when compared to simple RAG, is a notable advantage for users, as it often eliminates the need to verify answers directly from the source.
Instead, users can follow the reasoning in the derivation and identify faulty steps.
Figure~\ref{fig:incorrect-example} presents a real example of an incorrect derivation. In the application of the Refine rule, it is clear that the model has hallucinated facts not present in the hypotheses.

\vspace{-0.8cm}

\section{Conclusions}
\label{sec:conclusions}

In this paper we introduced Derivation Prompting, a new prompting technique inspired by logic derivations, to improve the generation step in the Retrieval-Augmented Generation framework for open-domain question answering.
Our experiments showed that Derivation Prompting significantly reduces the occurrence of unacceptable answers compared to traditional RAG and long-context window approaches.

However, the performance of Derivation Prompting is influenced by the size and capability of the underlying LLM.
While Claude Opus exhibited robust performance, smaller models like Claude Haiku showed a decrease in useful answers (though unacceptable answers were reduced), indicating the importance of model capacity when constructing effective derivations.

Future work will focus on refining Derivation Prompting by experimenting with different sets of rules and adjusting the number of initial hypotheses.
We are also working on formalizing the underlying formal language behind the application of the rules, and using this to add further verification methods to ensure the correctness of the resulting derivation.

We believe that this method can be applied to additional use cases and may be generalized to non-RAG scenarios with different sets of rules.
However, it is important to evaluate this method further to ensure its utility in such cases.

\bibliographystyle{splncs04}
\bibliography{custom}

\appendix

\section{Few-shot example}
\label{sec:few-shot}

\begin{scriptsize}
\noindent Hypotheses:
\noindent \textit{\{retrieved chunks removed for brevity, enumerated from 1 to 3\}}

\noindent User question: What is the meaning of a credit and how many hours of weekly study should be dedicated to a 13-credit course?

\noindent Extract | 2 | A credit is a measure of the dedication required for a subject. It is assumed that if a subject has more credits, it requires more hours of dedication. | Not a final answer

\noindent New hypothesis:
\noindent a. \textit{\{Previous conclusion is repeated, removed for brevity\}}

\noindent Extract | 2 | A credit approximately equals one hour of weekly study, throughout an entire semester. | Not a final answer

\noindent New hypothesis:
\noindent b. \textit{\{Previous conclusion is repeated, removed for brevity\}}

\noindent Instantiate | b | A credit approximately equals one hour of weekly study, throughout an entire semester. Therefore, a subject with 13 credits implies an approximate dedication of 13 hours per week. | Not a final answer

\noindent New hypothesis:
\noindent c. \textit{\{Previous conclusion is repeated, removed for brevity\}}

\noindent Concat | a,c | A credit is a measure of the dedication required for a subject. It is assumed that if a subject has more credits, it requires more hours of dedication. A credit approximately equals one hour of weekly study, throughout an entire semester. Therefore, a subject with 13 credits implies an approximate dedication of 13 hours per week. | Final answer

\end{scriptsize}

\end{document}